\documentclass[]{gmaart2022}

\usepackage[utf8]{inputenc} 
\usepackage[T1]{fontenc} 
 
\usepackage{amsmath,amssymb,amsfonts,mathtools,amsthm} 

\usepackage[babel,autostyle]{csquotes}

\usepackage{tabularx}
\newcolumntype{L}[1]{>{\raggedright\arraybackslash}p{#1}} 
\newcolumntype{C}[1]{>{\centering\arraybackslash}p{#1}} 
\newcolumntype{R}[1]{>{\raggedleft\arraybackslash}p{#1}} 
\usepackage{booktabs}

\usepackage{paralist}   

\usepackage[noend]{algpseudocode}

\algrenewcommand\algorithmicrequire{\textbf{Voraussetzung:}}
\algrenewcommand\algorithmicensure{\textbf{Abschlussbedingung:}}

\usepackage{listings} 
\usepackage{subcaption}  

\usepackage{epstopdf}
\usepackage{xcolor}
\selectcolormodel{gray}  

\usepackage{scrhack}
\usepackage{varwidth}
\usepackage{rotating} 
\usepackage[defaultlines=2,all]{nowidow}  
\usepackage{import}
\usepackage{placeins}
\usepackage{makecell}


\usepackage[version-1-compatibility]{siunitx} 

\usepackage{amsmath,amssymb,amsfonts,mathtools,amsthm} 

\usepackage{longtable} 
\usepackage{multirow}
\usepackage{tikz}
\usepackage{pgfplots}
\usepackage{smartdiagram}
\usetikzlibrary{patterns}
\usetikzlibrary{shapes} 
\usetikzlibrary{shapes.geometric, arrows,positioning}
\usepackage{smartdiagram}
\usesmartdiagramlibrary{additions} 
\AtBeginDocument{
}
\usepackage{xcolor}
\selectcolormodel{gray}

\usepackage[acronym]{glossaries} 
\usepackage{hyphenat}
\usepackage{etoolbox}
\usepackage{printlen}
\usepackage{graphicx}
\usetikzlibrary{calc, positioning, matrix}
\usepackage{standalone}

\newcommand{\newacr}[4][]{\newacronym[
	sort={\ifthenelse{\isempty{#1}}{#2}{#1}},
	]{#2}{#3}{#4}}
\glsdisablehyper

\robustify{\bfseries}

\usepackage{multirow}
\usepackage{array}

\newcolumntype{L}[1]{>{\raggedright\arraybackslash}p{#1}}
\newcolumntype{C}[1]{>{\centering\arraybackslash}p{#1}}
\makeglossaries
\newacronym{ai}{AI}{Artificial~Intelligence}
\newacronym[%
shortplural={ANNs},%
longplural={Artificial~Neural~Networks}%
]{ann}{ANN}{Artificial~Neural~Network}
\newacronym{api}{API}{Application~Programming~Interface}
\newacronym{clip}{CLIP}{Contrastive~Language-Image~Pre-training}
\newacronym[%
shortplural={FMs},%
longplural={Foundation~Models}%
]{fm}{FM}{Foundation~Model}
\newacronym{iou}{IoU}{Intersection~over~Union}
\newacronym[%
shortplural={LVLMs},%
longplural={Large~Vision~and~Language~Models}%
]{lvlm}{LVLM}{Large~Vision~and~Language~Model}
\newacronym{ml}{ML}{Machine Learning}
\newacronym{saa}{SAA+}{Segment~Any~Anomaly~+}
\newacronym{sam}{SAM}{Segment~Anything~Model}
\newacronym{sat}{SAT}{Scanning~Acoustic~Tomography}
\newacronym{sota}{SOTA}{State~Of~The~Art}
\newacronym{uav}{UAV}{Unmanned~Aerial~Vehicle}
\newacronym{vit}{ViT}{ Vision~Transformer}

\usepackage{doi}

\begin{document}


\hyphenpenalty=2000

\pagenumbering{roman}
\setcounter{page}{1}
\pagestyle{scrheadings}
\pagenumbering{arabic}

\setnowidow[2]
\setnoclub[2]

\renewcommand{\Title}{Are Foundation Models Ready for Industrial Defect Recognition? A Reality Check on
	Real-World Data}

\renewcommand{\Authors}{
	Simon~Baeuerle\textsuperscript{1,3,*},
	Pratik~Khanna\textsuperscript{1,3,*},
	Nils~Friederich\textsuperscript{1,2},
	Angelo~Jovin~Yamachui~Sitcheu\textsuperscript{1},
	Damir~Shakirov\textsuperscript{4},
	Andreas~Steimer\textsuperscript{4},
	Ralf~Mikut\textsuperscript{1}
}
\renewcommand{\Affiliations}{
	\textsuperscript{1}
	Institute for Automation and Applied Informatics (IAI)\\
	Karlsruhe Institute of Technology\\
	\textsuperscript{2}
	Institute of Biological and Chemical Systems (IBCS)\\
	Karlsruhe Institute of Technology\\
	\textsuperscript{3}
	Mobility Electronics, Robert Bosch GmbH\\
	\textsuperscript{4}
	Bosch Center for Artificial Intelligence, Robert Bosch GmbH\\
	*shared first, e-mail: simon.baeuerle@kit.edu
}


\renewcommand{\AuthorsTOC}{S.~Baeuerle, P.~Khanna, N.~Friederich, A.~J.~Yamachui~Sitcheu, D.~Shakirov, A.~Steimer, R.~Mikut} 
\renewcommand{\AffiliationsTOC}{Karlsruhe Institute of Technology, Robert Bosch GmbH} 

\setLanguageEnglish

\setupPaper 

\section*{Abstract}
\label{sec:abstract}
\glspl{fm} have shown impressive performance on various text and image processing tasks.
They can generalize across domains and datasets in a zero-shot setting.
This could make them suitable for automated quality inspection during series manufacturing, where various types of images are being evaluated for many different products.
Replacing tedious labeling tasks with a simple text prompt to describe anomalies and utilizing the same models across many products would save significant efforts during model setup and implementation.
This is a strong advantage over supervised \gls{ai} models, which are trained for individual applications and require labeled training data.
We test multiple recent \glspl{fm} on both custom real-world industrial image data and public image data.
We show that all of those models fail on our real-world data, while the very same models perform well on public benchmark datasets.\\


\section{Introduction}
\label{sec:introduction}
Defect pattern recognition is carried out during quality inspection in automotive series manufacturing.
During image-based quality inspection, images are recorded for a very wide range of manufacturing processes.
Beyond color images, advanced inspection technologies yield, e.g., depth images, x-ray images or \gls{sat} images.
The recording procedure is set up to capture defects that are specific to the respective manufacturing process.
This ensures high product quality and prevents defective products from being delivered to customers.
Sometimes, further expensive processing steps of a defective product can also be saved.
Due to the high volume of manufactured parts, an automated defect recognition procedure can significantly reduce manual inspection efforts.
Here, \glspl{ann} are increasingly used besides classic image processing techniques \cite{he_comprehensive_2025, bhatt_image-based_2021}.
However, a significant drawback of \gls{ann}-based classifiers is their reliance on large amounts of manually labeled training data. While this effort can be mitigated by unsupervised AI approaches~\cite{pleli_iterative_2024}, it remains a considerable hurdle. At the same time, simpler methods, such as a direct comparison to a reference image, are often insufficient. Real-world challenges like manufacturing tolerances, image registration errors, or varying brightness prevent such an approach from reliably detecting defects.

These limitations of both supervised models and simple heuristics motivate the exploration of a new class of powerful, pre-trained models. On various text and image processing tasks in other domains, \glspl{fm} have recently shown impressive performance \cite{bommasani_opportunities_2021, kirillov_segment_2023, radford_learning_2021}.
For example, the \gls{sam} and \gls{clip} generalize very well across many domains and datasets in a zero-shot setting, significantly reducing labeling efforts.
While \gls{clip} accepts small text inputs, \glspl{lvlm} like Gemini \cite{google_gemini_team_gemini_2025} are more powerful at the simultaneous processing of image and text input.
Domain experts can formulate text prompts without significant effort.
Promising ideas for prompting during quality inspection are, e.g., a description of the normal state of the product or a description of the visual or physical properties of the defects.
These reduced labeling efforts, combined with the additional opportunity to integrate domain expert knowledge via text input, would enable easier scaling across several products, i.e., with significantly lower efforts for each product.
The seemingly clear advantages over \gls{sota} approaches motivate the question of how suitable \glspl{fm} are for image-based quality inspection tasks.
To close these gaps for industrial use cases,  we analyze the applicability of various recent \glspl{fm} on real-world industrial data in this work.
The achieved performance is compared to the performance of the same models on a public dataset.

The main objective during quality inspection is the distinction of defective and defect-free products.
A classification model can perform such an inspection task with minimal setup.
However, it might not cover all desired functionalities fully:
In some cases, the classification of an \gls{ai} model is re-checked by a human operator.
Furthermore, a high level of explainability is preferred during model monitoring.
A model that outputs a full segmentation mask as opposed to only a single class makes both manual re-checking and model monitoring easier.
As such, we include both classification and segmentation models in our study.

The remainder of this work is structured as follows: The following section contains an overview of related work regarding e.g. \glspl{fm} and \gls{sota} pipelines for image segmentation.
The datasets that we use for benchmarking are introduced in Section~\ref{sec:datasets}.
The setup of our experiments is described in Section~\ref{sec:methodology} and respective results are shown in Section~\ref{sec:results}.
Section~\ref{sec:discussion} contains our interpretation of results and potential implications.\\


\section{Related Work} 
Several studies have investigated \glspl{fm} for computer vision. This section contains an overview of the most relevant models and studies in related applications.

Radford et al. \cite{radford_learning_2021} propose the \gls{clip} approach, which jointly trains an image encoder and a text encoder. It is trained on a dataset, which consists of 400 million pairs of images and corresponding text descriptions. \gls{clip} can perform zero-shot inference on unseen objects, i.e., it can be applied to new datasets without any finetuning. It outperforms a \gls{sota} supervised approach on multiple datasets \cite{radford_learning_2021}.

The \gls{sam}~\cite{kirillov_segment_2023} consists of an image encoder, a prompt encoder and a mask decoder. The image encoder uses a \gls{vit} ~\cite{dosovitskiy_image_2021}, which is pre-trained using the Masked Autoencoder approach ~\cite{he_masked_2022}. The prompt encoder encodes prompts like points, boxes, text or masks, whereas the prompt encoder for text uses the text encoder from \gls{clip}. The mask decoder creates masks based on the image embeddings and the prompt embeddings.
The \gls{sam} model is trained on the SA-1B dataset, which contains over one billion annotated masks on 11 million images. This covers a wide range of different objects, locations and scenarios. Images of people, buildings, vehicles, animals, and other elements from everyday life around the world are well represented.

Li et al. \cite{li_closer_2025}  propose \gls{clip}Surgery, which introduces small adaptations to the \gls{clip} architecture. They remove redundant features and modify the attention mechanism to link semantically similar regions better. This significantly improves the model's explainability.

Liu et al.~\cite{leonardis_grounding_2025} introduce GroundingDINO, which can detect objects based on text input. The feature enhancer includes cross-attention between text and image information. The language-guided query selection selects features that match the input text. The cross-modality decoder fuses the text modality with the image modality for the generation of the output regions. It is pretrained on large datasets. The prediction head outputs multiple object bounding boxes along with a text and a similarity score. A box threshold can be set to only include bounding boxes with a minimum score. A text threshold can be set to additionally filter for bounding boxes that match the given input text prompt well.

\glspl{lvlm} are models that accept multiple modalities such as text, images, audio and/or video as input.  They can be utilized in a wide range of applications. Gemini 2.5 Pro \cite{google_gemini_team_gemini_2025} can be directly applied to defect classification tasks, since it can process a text prompt along with an image. It has furthermore shown strong reasoning capabilities.

Zhang et al.~\cite{zhang_text2seg_2024} propose different pipelines of multiple foundation models, e.g., GroundingDINO + \gls{sam}, \gls{sam} + \gls{clip} or \gls{clip}Surgery + \gls{sam}. GroundingDINO and \gls{clip} can effectively capture semantic features and provide visual prompts for subsequent instance segmentation by \gls{sam}. Zhang et al. analyze the different pipelines on aerial images as taken by, e.g., an Unmanned Aerial Vehicle. 

Cao et al.~\cite{cao_segment_2023} propose the framework \gls{saa} for zero-shot anomaly segmentation. This includes GroundingDINO to propose abnormal regions which are then fed into a second model, such as \gls{sam} to refine the abnormal regions. They specifically study defect detection and analyze performance on the public industrial dataset \textit{MVTec AD}. 

Xu et al.~\cite{xu_customizing_2025} combine human expert knowledge with the capabilities of Visual-Language-Foundation  Models. They study different prompts for the task of anomaly detection in images. This includes simple prompts that only query for any defect or anomaly and more advanced prompts that provide, e.g., more detailed information about the shown object or the expected defects.

The Text2Seg approach contains multiple of the most relevant \gls{fm} models.
They have been successfully tested on an image segmentation task.
SAA+ follows a similar approach to Text2Seg and its authors even include the task of defect recognition into their study.
As outlined above, the Gemini model is widely applicable and has shown promising capabilities.
This raises high expectations towards the application of those models during industrial defect recognition.
While there is a wide range of further models that would also be interesting in this context, we limit the scope of our study to these models.



\section{Datasets}
\label{sec:datasets}
During this research, we utilized three different image datasets: our custom real-world industrial dataset \textit{IndustrialSAT}, the public industrial dataset \textit{MVTec AD} and the general public image dataset \textit{Oxford-IIIT-Pet}. The main focus of our study is the real-world data. The public datasets serve as a reference.

\subsection{Oxford-IIIT-Pet}

\begin{figure}[!htb]
	\centerline{\includegraphics[width=.9\columnwidth]{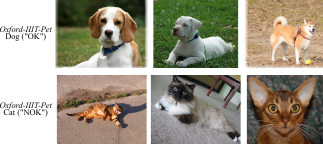}}
	\caption{Exemplary images from the dataset \textit{Oxford-IIIT-Pet} \cite{parkhi_cats_2012}}
	\label{fig:dataset_oxford}
\end{figure}

The dataset \textit{Oxford-IIIT-Pet}~\cite{parkhi_cats_2012} contains images of various pet animals in different everyday scenarios.
The animals are recorded both outdoors and indoors in different scales, pose and lighting.
Such animal images are found across the internet and are well-represented in the SA-1B dataset.
During this work, only cats and dogs are utilized.
The dog images are treated as defect-free images ("OK"), whereas cat images are treated as defective ("NOK").
During testing, we utilize 140 dog images and 60 cat images.
Examples are shown in Figure~\ref{fig:dataset_oxford}.

\subsection{MVTec AD}

\begin{figure}[!htb]
	\centerline{\includegraphics[width=.9\columnwidth]{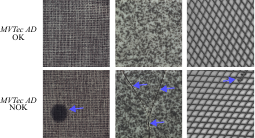}}
	\caption{Exemplary images from the dataset \textit{MVTec AD} \cite{bergmann_mvtec_2019} from the classes carpet (left), tile (center) and grid (right). Top row: defect-free images. Bottom row: defective images.}
	\label{fig:dataset_mvtec}
\end{figure}

The \textit{MVTec AD} dataset~\cite{bergmann_mvtec_2019} is utilized during the benchmarking of various models for industrial anomaly detection~\cite{batzner_efficientad_2024, del_bimbo_padim_2021}.
In this work, we analyze only object categories that are visually related to our domain of electronic packaging.
This includes the categories carpet, grid, leather, tile and wood with a resulting test dataset of 253 defect-free images ("OK") and 76 defective images ("NOK").
No distinction is made between different defect types.
Examples from the categories carpet, tile and grid are shown in Figure~\ref{fig:dataset_mvtec}.

\subsection{IndustrialSAT}

\begin{figure}[!htb]
	\centerline{\includegraphics[width=.9\columnwidth]{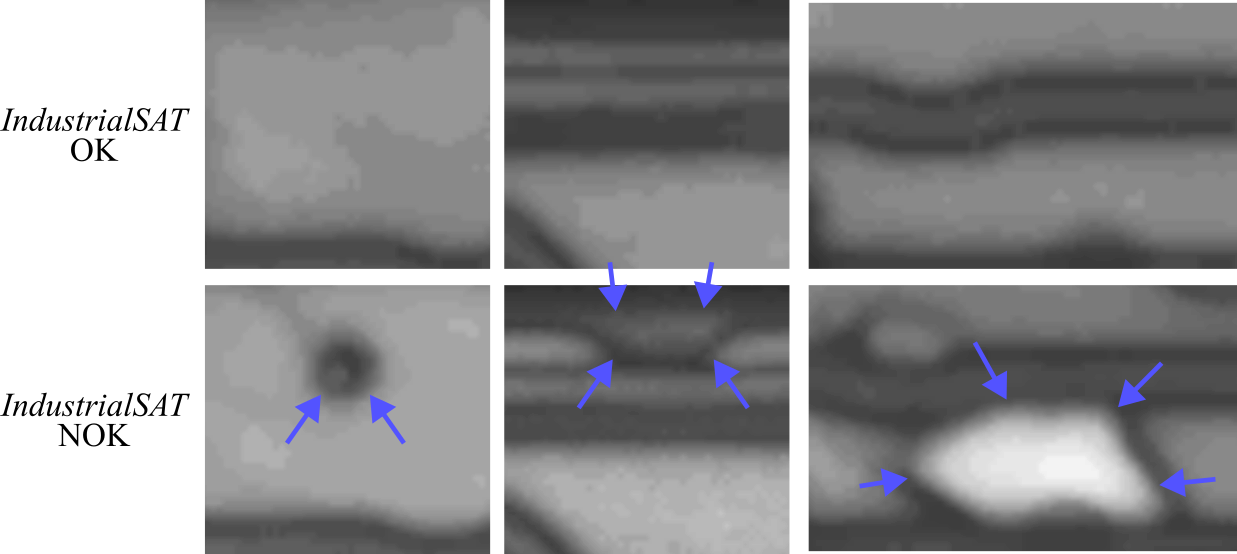}}
	\caption{Exemplary images from the dataset \textit{IndustrialSAT}. Crop of defective images (bottom row) and corresponding regions without defect (top row).}
	\label{fig:dataset_IndustrialSAT}
\end{figure}

Our custom real-world industrial dataset \textit{IndustrialSAT} is a dataset of greyscale images.
They are recorded with \gls{sat} \cite{liu_using_2017, yu_scanning_2020} as follows.
The electronic package is submerged in water.
A toolhead with an ultrasonic sender and receiver is moved to a position above the package.
An ultrasonic wave is sent out and the signal reflection is recorded.
This yields a time-series of data points for this position.
From the time-series, the signal value at a certain time value is extracted with special postprocessing methods.
This outputs a single value for this position of the toolhead.
The procedure is repeated for multiple toolhead positions.
The toolhead positions correspond to the pixels of the resulting image.
The extracted signal values correspond to the grey values of the resulting image.
\gls{sat} is used for quality inspection during electronic packaging of products in the field of, e.g., sensors, electronic control units or power electronics.
These products are highly relevant for electric and autonomous vehicles.
Examples of defect types that could occur during electronic packaging are cracks, voids or delaminations between or within the different layers of an electronic package.
Our test dataset, \textit{IndustrialSAT}, contains 231 defect-free images and 32 images with one of the defect types: void, crack or delamination.
Exemplary defects on \gls{sat} images are shown in the bottom row of Figure~\ref{fig:dataset_IndustrialSAT}.
The shown images are cropped down to the immediate area around the defects.
The top row of Figure~\ref{fig:dataset_IndustrialSAT} shows the same crop position of a non-defective product.
To evaluate our model, we assigned the defective label to images with defects of any size.
No distinction is made with respect to different defect types during model evaluation.

During a preliminary study, we trained a classification model on the dataset \textit{IndustrialSAT}.
This was built by concatenating a pre-trained feature extractor based on a ResNet model~\cite{he_deep_2015} with a multi-layer perceptron.
This classifier reached an F2-score of 0.82 on a stratified five-fold cross-validation dataset split.
This proves that the defects on this dataset (such as visualized in Figure~\ref{fig:dataset_IndustrialSAT}) can be recognized by a \gls{ml} model.

The \gls{sat} images during manufacturing look similar to each other at first sight.
During another preliminary study, we tested a simple approach: we aligned all images with respect to a predefined reference image.
Then, we calculated the difference on the pixel level with respect to the reference image.
However, this was not sufficient to detect defects such as voids or delaminations.
The variation during the SAT recording, along with the variation in mechanical tolerances, was too large.


\section{Methodology} 
\label{sec:methodology}
We test multiple \glspl{fm} both on the task of segmentation and the task of classification on the above datasets.
Our experiments are set up as follows.

\subsection{Segmentation}
\label{segmentation}
For the task of segmentation, we use the \gls{iou} as a metric to assess the segmentation output:
\begin{equation}
	IoU = \frac{|A \cap B|}{|A \cup B|},
\end{equation}
with $A$ being the set of pixels predicted positive by the model and $B$ being the set of pixels marked positive in the ground-truth.
The \gls{iou} metric yields values between zero and one, whereas one corresponds to an optimum result.
This metric penalizes especially the slip of small defects (=false negatives): The correctly detected defective area (intersection of defective prediction and ground truth) is divided by the union of the defective prediction and the ground truth.
A miss of a small defect, such as a void, will thus return low metric values.
Such false negatives are especially critical since defective products would be delivered to customers.

Three pipelines are chosen from Text2Seg for evaluation: GroundingDINO + \gls{sam}, \gls{sam} + \gls{clip} and \gls{clip}Surgery + \gls{sam}.
The Git repository of the \gls{clip}Surgery paper is used both for the \gls{clip} Surgery model and the \gls{clip} model.
The original Git repository by IDEA research \cite{idea-research_GitHub_2025} is used for the GroundingDINO model.
All of our Text2Seg pipelines use the huge backbone of \gls{sam}.
The base \gls{vit} is used for both \gls{clip} and \gls{clip} Surgery with weights \textit{CS-ViT-B/16} \cite{xmed-lab_GitHub_2025}  for \gls{clip} Surgery and weights \textit{ViT-B/16} \cite{openai_GitHub_2025} for \gls{clip}. 
The checkpoint \textit{groundingdino swint ogc} \cite{idea-research_GitHub_2025} is used for the GroundingDINO model, along with a box threshold of 0.35 and a text threshold of 0.25.
Minor issues such as the handling of empty segmentation masks, are resolved to enable an automated end-to-end evaluation in all cases.
The utilized text prompts are short and simple, e.g., “defect” or “cat”.

For \gls{saa}, we use GroundingDINO with weights \textit{groundingdino swint ogc} as the region proposal network and a \gls{sam} model with the huge backbone as region refiner.
The box threshold is set to 0.1 and the text threshold to 0.1.
While the authors of Text2Seg evaluate their approach on the task of semantic segmentation in remote sensing images as taken, e.g., by an \gls{uav}, the authors of \gls{saa} focus specifically on the segmentation of anomalies and also utilize the \textit{MVTec AD} dataset during evaluation.

Since initial tests for the segmentation models have shown insufficient performance on both the real-world \textit{IndustrialSAT} dataset and the public industrial \textit{MVTec AD} dataset,
we extend the evaluation for those models to include the more general public dataset \textit{Oxford-IIIT-Pet}.
This is done to validate the correct setup of our software pipelines and to gain additional insights.

The chosen model pipelines represent a diverse setup of multiple prominent \gls{sota} \glspl{fm}.
The models are all executed in their inference mode and are tested on the defective images of all three datasets that were introduced in the previous section.
The reported \gls{iou} metric values are averaged over all defective images.
This yields an insight into how well defects can be recognized in the different datasets.\\

\subsection{Classification}
\label{classification}
For the task of classification, we analyze the \gls{lvlm} model Gemini 2.5 Pro.
It features a high token limit, which enables long prompts.
Furthermore, it can process multi-modal input, i.e., it can receive prompts consisting of both written text and images.
The model's output includes a single class label and a corresponding reasoning string.
The reasoning string is more verbose and gives a deeper insight into how the model has made a decision, whereas the class label can be used during automated postprocessing.
For each dataset, a reference image is defined.
This reference is used to automatically align all images, ensuring the same position and orientation of the product in all images.

The testing is carried out using the Google Cloud Platform, which offers a convenient access to the model via an \gls{api}.
It is carried out using both defect-free and defective images from the industrial datasets \textit{IndustrialSAT} and \textit{MVTec AD}.
Initial experiments have shown promising performance on industrial data.
Thus, an assessment on the general public dataset \textit{Oxford-IIIT-Pet} is omitted, since the aim of this paper is to assess model performance on real-world industrial data.

The F2-score is used as a metric during evaluation:
\begin{equation}
	F_{2} = \frac{5 \cdot \text{precision} \cdot \text{recall}}{4 \cdot \text{precision} + \text{recall}}
\end{equation}
Precision is the ratio of true positives (defects detected as defects) divided by the sum of true and false positives.
Recall is the ratio of true positives divided by the sum of true positives and false negatives.
As compared to F1-score, the F2-score puts a higher weight on recall than on precision.
This penalizes false negatives (slips of defective products) more heavily compared to false positives (wrong alerts for defect-free products)~\cite{hashemi_asymmetric_2019}.
This is desired, since it is better to re-check a suspicious product then to deliver a defective product to the customer.

When querying the Gemini model, a test image with an unknown state is sent to the model along with a prompt as described below.
The prompt may include a predefined reference image.
Two types of prompts were defined for the Gemini model.
The basic prompt consists of a) a simple text prompt that queries for any anomalies or defects and b) an exemplary defect-free image.
The refined prompt additionally includes specific information about the physical structure of e.g. the electronic package in \textit{IndustrialSAT} or the object properties in \textit{MVTec AD} and information about the visual appearance of the respective defects.
This follows the conceptual approach of varying information depth in prompts by Xu et al.~\cite{xu_customizing_2025}.
On \textit{MVTec AD}, the prompts are as follows.
The basic prompt for \textit{MVTec AD} is “Please determine whether the image contains anomalies or defects. If yes, give a specific reason” - similar to the naive prompt studied by Xu et al.~\cite{xu_customizing_2025}.
The refined prompt for \textit{MVTec AD} is “Please determine whether the last image given about object contains anomalies or defects.
If so, please provide a specific reason.
Normally, the image given should depict a clear and identifiable object.
It may have defects such as broken/bented parts, contaminations, threads, color stains, cuts, holes, scratches, liquids, glue, folds, pokes, oil or glue strips.”.
The formulation of the prompts for \textit{IndustrialSAT} is made in a close alignment with process engineers, who are responsible for the respective packaging processes.
These prompts include specifics of the respective product and thus may not be published in full detail.


\section{Results}
\label{sec:results}
This section contains the benchmarking results of both the segmentation and classification models on the different datasets as introduced before.

\subsection{Segmentation}
\begin{figure}[!htb]
	\centerline{\includegraphics[width=1\columnwidth]{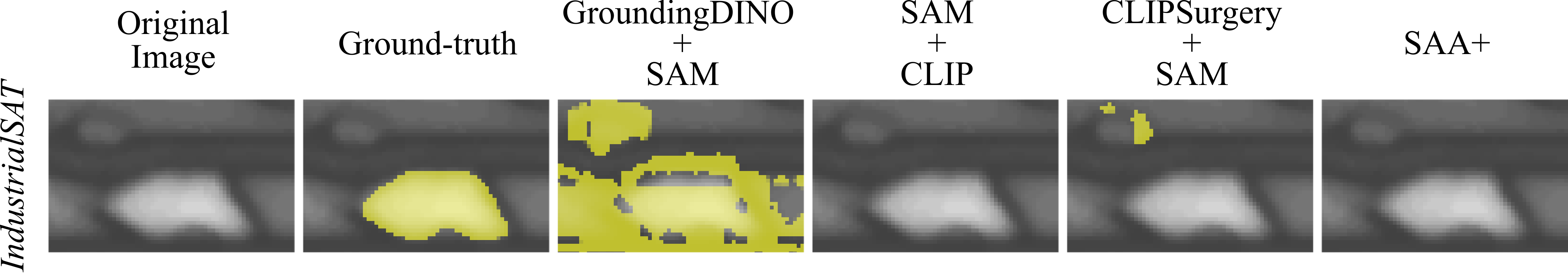}}
	\caption{Exemplary segmentation results on dataset \textit{IndustrialSAT}.
		A yellow color overlay is used to indicate defective regions as defined in ground-truth data or as predicted by the respective model.
		All models fail to detect the defects in most images.}
	\label{fig:results_IndustrialSAT}
\end{figure}

\begin{figure}[!htb]
	\centerline{\includegraphics[width=1\columnwidth]{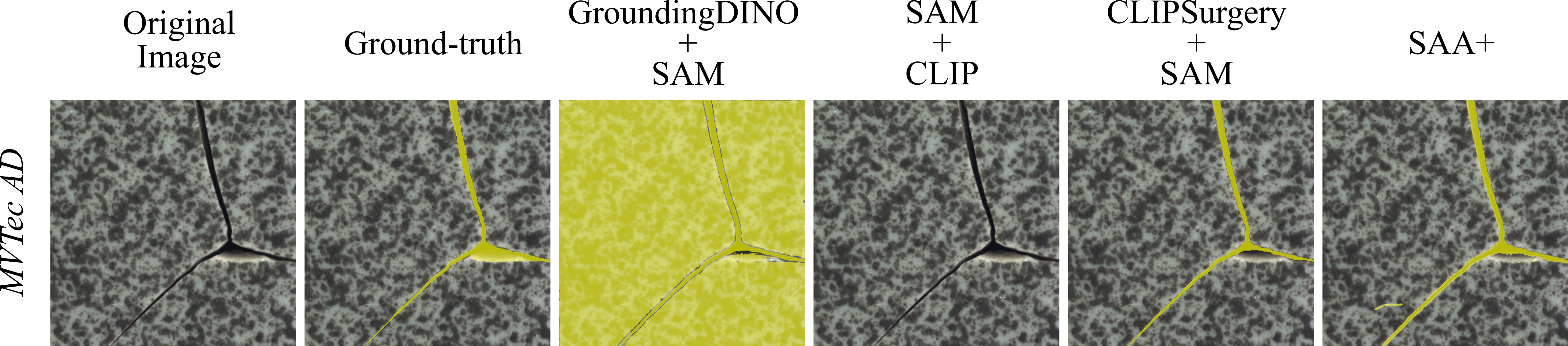}}
	\caption{Exemplary segmentation results on dataset \textit{MVTec AD} \cite{bergmann_mvtec_2019}.
		A yellow color overlay is used to indicate defective regions as defined in ground-truth data or as predicted by the respective model.
		In some cases, defects can be accurately recognized.
		In other cases, the models fail to recognize the defects.}
	\label{fig:results_mvtec}
\end{figure}

\begin{figure}[!htb]
	\centerline{\includegraphics[width=1\columnwidth]{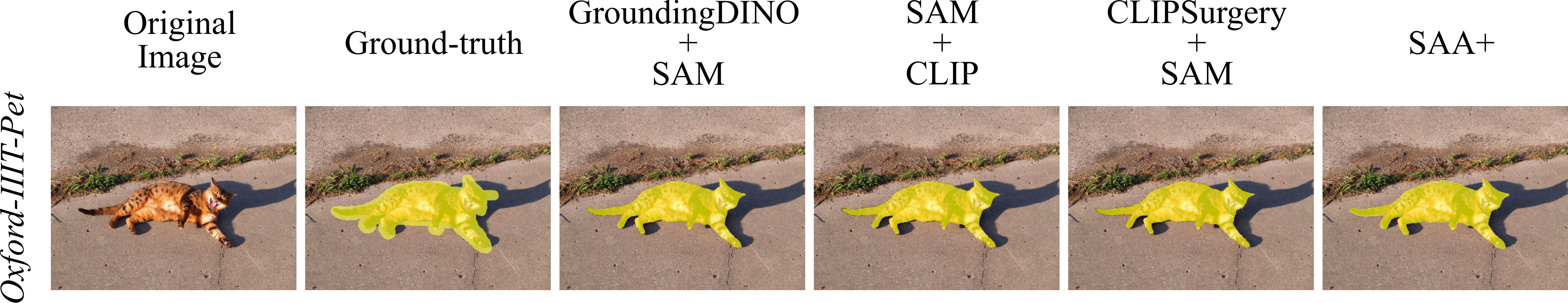}}
	\caption{Exemplary segmentation results on dataset \textit{Oxford-IIIT-Pet} \cite{parkhi_cats_2012}.
		A yellow color overlay is used to indicate defective regions as defined in ground-truth data or as predicted by the respective model.
		Animals such as the cat shown in this image can be segmented very accurately by all tested models.}
	\label{fig:results_oxford}
\end{figure}

The \gls{iou} metric values for the segmentation models are reported in Table~\ref{tab:segmentation_results}.
Essentially, the defects in the \gls{sat} images cannot be detected in any of the tested pipelines.
Visual examples are shown in Figure~\ref{fig:results_IndustrialSAT} for each model to give deeper insights.
The pipeline based on GroundingDINO + \gls{sam} seems to be very sensitive and segments large defect-free regions of the image.
For the other pipelines, the defects are missed entirely.
In some cases, the pipeline \gls{clip}Surgery + \gls{sam} segments geometric features of sub-components instead of defective regions.
For the \textit{MVTec AD} dataset, metric values are rather low.
A visual depiction of results in Figure~\ref{fig:results_mvtec} shows that some defects can be detected accurately.
For the public dataset \textit{Oxford-IIIT-Pet}, the \gls{iou} metric results are promising, with \gls{iou} scores ranging from 0.65 to 0.80.
A visual depiction of results in Figure~\ref{fig:results_oxford} shows that the cat can be segmented very well by all models.
We conducted a thorough inspection of various images and report that the model's segmentation results align with the depicted cat even more closely than the ground truth masks.
Since the cat is configured to represent defective regions, this validates the correct setup of all pipelines for the segmentation task.
Furthermore, we report a correct segmentation of other animals (e.g. dogs).
Only \gls{iou} scores larger than 0.60 are considered for a comparison between models (indicated in Table~\ref{tab:segmentation_results} by bold print), since lower values are too low to offer practical value.

\begin{table}[htbp]
	\centering
	\caption{\gls{iou} scores on all datasets for various \glspl{fm} pipelines.
		None of the methods can detect defects on our internal dataset \textit{IndustrialSAT}.
		Only \gls{saa} works somehow reasonable on the public \textit{MVTec AD} dataset.
		All segmentation models work well on the public dataset \textit{Oxford-IIIT-Pet}.}
	\label{tab:segmentation_results}
	\begin{tabular}{|L{0.27\linewidth}|C{0.23\linewidth}|C{0.08\linewidth}|C{0.18\linewidth}|C{0.07\linewidth}|}
		\hline
		\multirow{3}{*}{\textbf{Classes and datasets}} &
		\multirow{3}{*}{\shortstack{\textbf{GroundingDINO}\\\textbf{+}\\\textbf{\gls{sam}}}} &
		\multirow{3}{*}{\shortstack{\textbf{\gls{sam}}\\\textbf{+}\\\textbf{\gls{clip}}}} &
		\multirow{3}{*}{\shortstack{\textbf{\gls{clip}Surgery}\\\textbf{+}\\\textbf{\gls{sam}}}} &
		\multirow{3}{*}{\textbf{\gls{saa}}} \\
		& & & & \\ 
		& & & & \\ \hline
		Defect types void, crack and delamination in dataset \textit{IndustrialSAT} & 0.00 & 0.00 & 0.00 & 0.00 \\ \hline
		Various defects such as cracks, holes, scratches in dataset \textit{MVTec AD} & 0.13 & 0.05 & 0.19 & 0.52 \\ \hline
		Animal cat in dataset \textit{Oxford-IIIT-Pet} & \textbf{0.80} & 0.65 & 0.78 & 0.77 \\ \hline
	\end{tabular}
\end{table}

\subsection{Classification}

Results for the classification use case are shown in Table~\ref{tab:classification_results}.
The F2-score values are far too low to be suitable for industrial applications during quality inspection.
However, the model can recognize some defects accurately, both with the basic and refined prompt strategy.
Also, we report that the reasoning string output looks promising and shows that the \gls{lvlm} model was able to recognize the package build-up to a certain extent.

The Gemini model performs significantly better on the public industrial dataset \textit{MVTec AD} as compared to our internal industrial dataset \textit{IndustrialSAT}.
Interestingly, the highest metric results were achieved with simple prompts, which consist of a naive text prompt for defects along with a defect-free reference image.
The maximum F2-score of 0.37 achieved by the Gemini model on \textit{IndustrialSAT} is far below the F2-score of 0.82 reached with a supervised \gls{ai} model in a preliminary study.

In analogy to the \textit{Oxford-IIIT-Pet} dataset on the segmentation task, the good performance on the \textit{MVTec AD} dataset validates the correct setup of our evaluation pipeline for the classification task.

\begin{table}[htbp]
	\centering
	\caption{Classification results (F2-score) on both industrial datasets for the Gemini 2.5 Pro \gls{lvlm} model.
		While the model performs well on the \textit{MVTec AD} dataset, its performance on \gls{sat} imaging data is insufficient for industrial-quality inspection.}
	\label{tab:classification_results}
	\begin{tabular}{|l|c|c|}
		\hline
		\multirow{2}{*}{\textbf{Dataset}} &
		\multirow{2}{*}{\shortstack{\textbf{Gemini}\\(basic prompt)}} &
		\multirow{2}{*}{\shortstack{\textbf{Gemini}\\(refined prompt)}} \\
		& & \\ \hline
		\textit{IndustrialSAT} & 0.37 & 0.31 \\ \hline
		\textit{MVTec AD}      & 0.99 & 0.92 \\ \hline
	\end{tabular}
\end{table}


\section{Discussion}
\label{sec:discussion}
The three tested datasets can be ordered according to their visual similarity to datasets like SA-1B that are used to train the \glspl{fm}:
\textit{Oxford-IIIT-Pet} is in a very similar visual domain as SA-1B.
\textit{MVTec AD} is more focused on defects and includes industrially relevant parts, but some of its object categories, such as carpets and wood, are likely to be found in, e.g., SA-1B.
However, our real-world dataset, \textit{IndustrialSAT}, differs significantly from datasets like SA-1 B.
It is not a color image, but rather a greyscale image.
The geometric features that are shown resemble mostly geometric primitives, such as rectangular areas and line features.
The performance of all the \glspl{fm} is decreasing along with the visual similarity of the test dataset to training datasets such as SA-1B.
The domain gap seems to be significant, especially for the real-world dataset \textit{IndustrialSAT}.
This is likely to be the cause of the insufficient performance on our real-world industrial data.

Furthermore, the performance difference on the datasets \textit{MVTec AD} and \textit{Industrial\gls{sat}} motivates the question of how well \textit{MVTec AD} can represent advanced imaging procedures such as \gls{sat}.
The \textit{MVTec AD} dataset contains industrially relevant defects, but it is in a similar visual domain as public datasets.
For example, screws and fabric-like materials can be found in everyday items, which are likely to be included in the training datasets of \glspl{fm} to a certain extent.
Defect patterns such as voids and cracks in \gls{sat} imaging are not well represented.

The reasoning strings of the Gemini model show a deeper understanding of the product build-up.
This may not be fully correct, but it gives an interesting starting point for further studies.
If this knowledge is available at least partially for the model, more advanced prompt strategies could leverage this knowledge better.
The Gemini model was the only model that could detect some of the defects and a more refined prompting strategy seems to be a promising way to improve performance.\\ 


\section{Conclusion and outlook}
\label{sec:conclusion_and_outlook}
Industrial quality inspection regularly involves advanced imaging procedures such \gls{sat}.
In this work, we investigate the use of \glspl{fm} to improve industrial defect recognition.
None of the tested \glspl{fm} were able to detect defects in our real-world \gls{sat} images.
The results on other public datasets validate the correct setup of our models.
While an application of \glspl{fm} to such data could be scaled across many manufacturing stations, currently available \glspl{fm} are not ready for practical application in series manufacturing for advanced inspection technologies such as \gls{sat} imaging.

The low performance in \textit{IndustrialSAT} may be due to a large domain gap between the training data of \glspl{fm}.
Future work could thus involve fine-tuning \glspl{fm} on image data that better matches the real-world industrial domain.
Furthermore, more advanced prompt strategies for \glspl{lvlm} seem to be a promising direction for an in-depth follow-up study.

\section{Acknowledgments}
The Helmholtz Association funds this project under the research school "Helm\-holtz Information and Data Science School for Health (HIDSS4Health)",
the program "Natural, Artificial and Cognitive Information Processing (NACIP)" and the Initiative and Networking Fund on the HAICORE@KIT partition.
This work was funded by the European Union (NextGenerationEU) and the German Federal Ministry for Economic Affairs and Energy (BMWE) based on a decision by the German Bundestag.
The authors acknowledge ﬁnancial support within the SmartMan project (reference number: 13IK033) in the Kopa35c program.
\\

We describe the contributions of Simon Baeuerle (SB), Pratik Khanna (PK), Nils Friederich (NF), Angelo Jovin Yamachui Sitcheu (AJYS), Damir Shakirov (DS), Andreas Steimer (AS) and Ralf Mikut (RM) according to CRediT \cite{brand_beyond_2015}:
Writing-Original Draft: SB;
Writing-Review \& Editing: PK, NF, AJYS, DS, AS, RM;
Conceptualization: SB, RM; Investigation: PK (within masters thesis \cite{khanna_foundation_2025});
Methodology: SB, PK, NF, AJYS, DS, AS, RM; Software: PK;
Supervision: SB, RM;
Project administration: SB, RM;
Funding Acquisition: SB, RM.



\addtocontents{toc}{\protect\newpage}



\end{document}